\documentclass[conference]{IEEEtran}
\IEEEoverridecommandlockouts

\usepackage[utf8]{inputenc}
\usepackage[english]{babel}
\usepackage[T1]{fontenc}
\usepackage{tikz}
\usepackage{tabularx}
\usepackage{cite}
\usepackage{amsmath,amssymb,amsfonts}
\usepackage{algorithmic}
\usepackage{graphicx}
\usepackage{textcomp}
\usepackage{xcolor}
\usepackage{lipsum}

\usepackage{hyperref}
\hypersetup{
  colorlinks   = true, %Colours links instead of ugly boxes
  urlcolor     = blue, %Colour for external hyperlinks
  linkcolor    = red, %Colour of internal links
  citecolor    = green %Colour of citations, could be ``red''
}

\newcommand\copyrighttext{%
  \footnotesize  © 2021 IEEE.  Personal use of this material is permitted.  Permission from IEEE must be obtained for all other uses, in any current or future media, including reprinting/republishing this material for advertising or promotional purposes, creating new collective works, for resale or redistribution to servers or lists, or reuse of any copyrighted component of this work in other works. 
 Original DOI: \url{https://doi.org/10.1109/FUZZ45933.2021.9494444}
 }
\newcommand\copyrightnotice{%
\begin{tikzpicture}[remember picture,overlay]
\node[anchor=south,yshift=20pt] at (current page.south) {\fbox{\parbox{\dimexpr\textwidth-\fboxsep-\fboxrule\relax}{\copyrighttext}}};
\end{tikzpicture}%
}
    
\begin{document}

\bstctlcite{IEEEexample:BSTcontrol}
\title{Fuzzy Temporal Protoforms for the Quantitative Description of Processes in Natural Language 
 \thanks{This research was funded by the Spanish Ministry for Science, Innovation and Universities, the Galician Ministry of Education, University and Professional Training and the ERDF/FEDER program  (grants TIN2017-84796-C2-1-R, ED431C2018/29 and ED431G2019/04). 
 }
}

\author{\IEEEauthorblockN{Yago Fontenla-Seco, Alberto Bugarín, Manuel Lama}
 \IEEEauthorblockA{\textit{Research Center on Intelligent Technologies (CiTIUS)} \\
  \textit{University of Santiago de Compostela}\\
  Santiago de Compostela, Spain \\
  \{yago.fontenla.seco, alberto.bugarin.diz, manuel.lama\}@usc.es}
}

\maketitle

\begin{abstract}
In this paper, we propose a series of fuzzy temporal protoforms in the framework of the automatic generation of quantitative and qualitative natural language descriptions of processes. The model includes temporal and causal information from processes and attributes, quantifies attributes in time during the process life-span and recalls causal relations and temporal distances between events, among other features. Through integrating process mining techniques and fuzzy sets within the usual Data-to-Text architecture, our framework is able to extract relevant quantitative temporal as well as structural information from a process and describe it in natural language involving uncertain terms. A real use-case in the cardiology domain is presented, showing the potential of our model for providing natural language explanations addressed to domain experts.
\end{abstract}


\vspace*{-0.45cm}

\copyrightnotice

%!TEX root = main.tex

\section{Introduction}\label{section:introduction}
Processes constitute a useful way of representing and structuring the activities and resources involved in organization Information Systems (IS) from almost any domain. 
Daily, more event data is being produced and recorded, making it necessary to provide organizations with tools capable of processing such vast amounts of data and extracting the valuable knowledge hidden in it.

Process data is recorded as event logs, but its behavior is usually represented as process models (in a variety of notations \cite{Aalst16}) that represent in a graphical manner the activities that take place in a process as well as the dependencies between them. Other relevant properties of the process tend to be included in the model such as temporal properties, process execution-related statistics, etc.
Apart from process models, information about these properties is conveyed to users through visual analytics, as they are commonly used when providing advanced analytics \cite{Aalst16}.
However, in real life scenarios process models are very complex, with a high number of relations between activities.
Furthermore, the amount of information that can be added to enhance the process model is very high, and the visual analytics related to said process information are quite difficult to be understood and related to the underlying process for users, as deep knowledge of process modeling and analytics is required.

%Parallel to visual analytics, 
In the Natural Language Generation (NLG)\cite{Reiter2000} and Linguistic Descriptions of Data (LDD)\cite{RAMOSSOTO2016} fields, different methods for generating insights on data through natural language have been under development. Through different techniques, they aim to provide users with natural language texts that capture or summarize the most characteristic aspects of some data. This information can be easily consumed by users, as \textit{i)} natural language is the inherent way of communicating for humans, therefore it does not rely on their capabilities to identify or understand patterns, trends, etc. from visual representations; and \textit{ii)} it may include uncertain terms or expressions, which are very effective for communication. In this sense, research suggests that in some domains knowledge and expertise are required to understand graphical information \cite{petre95} and proves that domain experts can take better decisions based on textual descriptions than on graphical displays \cite{Law2005}. Therefore, natural language descriptions seem a good approach to enable or enhance the understanding  of processes and its analytics as they can summarize, combine and communicate information in ways it would not be possible with visual representations.

In this paper, we investigate a real-life use case of a process in the health-care domain which could potentially benefit from natural language descriptions in order to achieve a better understanding of what is really happening in it.

We propose a series of fuzzy temporal protoforms (fuzzy linguistic descriptions of data) in the framework of the automatic generation of quantitative and qualitative natural language descriptions of processes.
With a general model that includes temporal and causal information from processes and its attributes we are able to 
recall causal relations and temporal distances between events, among other features.
The use of fuzzy linguistic descriptions of data allows for modeling and managing the inherent imprecision of linguistic terms, which is very useful when summarizing temporal and other data. 
By introducing imprecision in descriptions related to frequency and temporal characteristics of processes the expressiveness of the approximation is enhanced.
As fuzzy linguistic variables represent a language abstraction that compacts information and relations about sets of data, fuzzy quantified statements provide a more human-friendly interface than process models or visualization techniques.
This approach also introduces the description of causal and temporal relationships between activities of a process, including both frequency and temporal characteristics.

%The following sections are structured as follows. 
Section \ref{section:state-of-art} gives a deeper background in NLG, LDD and its applications on process data and event logs. It also contains basic concepts of fuzzy quantified statements and process mining used in the proposed solution.
Section \ref{section:proposal} introduces the proposed protoforms and an overview of the generation process. 
Section \ref{sec:conclusions} presents an evaluation of the proposal and some concluding remarks.
%!TEX root = main.tex

\section{Background and Related work}\label{section:state-of-art}
The generation of natural language texts from data is a task which originated within the NLG field \cite{Reiter2000}.
Particularly, the generation of natural language descriptions over data has been traditionally a task tackled by the Data-to-Text (D2T) research community \cite{Reiter2007}.
Parallel to NLG and D2T systems, in the fuzzy logic realm, the paradigms of Computing with Words and Linguistic Descriptions (or summaries) of Data (LDD) emerge for modelling and managing uncertainty in natural language with the use of fuzzy sets \cite{Yager1982}.
These paradigms use the concepts of linguistic descriptions of data and protoform
\cite{Zadeh2002}, which aim on providing summaries involving linguistic terms with some degree of uncertainty or ambiguity present on them.

\subsection{Linguistic Descriptions of Data}

Linguistic summaries or descriptions of data have been investigated by many researchers and applied in multiple domains. Classically, fuzzy quantified sentences of type-I and type-II have been the most used ones in the literature since their inception in early 1980's \cite{Yager1982}.
From generating weather forecasts \cite{RamossotoWeather} or data-base summarization \cite{Kacprzyk2002} to temporal series summarization \cite{Castillo-OrtegaMS11}.
However, they have been only investigated succinctly on process data \cite{Dijkman17, Wilbik15}.

% Describir linguistic summaries y sus componentes
Linguistic descriptions of data are understood as sets of instanced fuzzy quantified statements that are computed according to a dataset and a knowledge base for a given application domain for summarizing knowledge about variables and their values \cite{Ramos-Soto21}.
Fuzzy quantified statements follow predefined formal structures or templates, that are referred to as protoforms which are composed mainly of four aspects:

\begin{itemize}
 \item A referential $X$ is a set of objects for which certain property or set of properties holds (e.g. the set of cases from an event log).
 \item A summarizer $A$ used to indicate some property or  aggregation of properties (e.g. "long waiting time" or "long waiting time and high number of medical tests") of the object or referential of interest.
 \item A (fuzzy) quantifier $Q$ (e.g. "several") used to express the quantity or proportion of data from the referential which fulfills the properties indicated by the summarizer.
 \item The degree of truth $T$ used to relate the validity of the protoform. 
 Instancing a protoform involves assigning values to its elements (referential, quantifier and summarizer) and  computing its truth degree. 
 The truth degree can be calculated using any valid quantification model 
 \cite{DanielSanchez2013, Andrea20}.
\end{itemize}

Combining these elements, a sentence like "In several cases there was a long waiting time between the Medical Surgical (MS) session of a patient and its surgery" can be created from type-I protoform:
\begin{equation}   \label{eq:1}
   Q \ X's \ are \ A
\end{equation}

In some cases, one may want a finer-grained description. 
A qualifier can be added to the description to better define the scope of the sentence, giving place to type-II protoforms.
\begin{itemize}
 \item A qualifier $B$, can make reference to any property or aggregation of properties of the referential. It defines a subset of it which fulfills a property to a certain degree and will be evaluated against summarizer and quantifier.
\end{itemize}
Sentences like "In most cases where patients were males, there was a long waiting time between the MS session of a patient and its surgery" can be created from a type-II protoform:
\begin{equation} \label{eq:2}
    Q \ BX's \ are \ A
\end{equation}

Both summarizer, qualifier and quantifier take the form of a linguistic variable.
Linguistic variables model the partitioning of the domain of a numeric or categorical variable into several properties (e.g., waiting time = \{\textit{really short}, \textit{short}, \textit{as expected}, \textit{long}, \textit{extremely long}\}), where each property is known as a linguistic value and is associated to a membership function that measures the degree in which different values of the original variable fulfill that property.
These membership functions are usually represented as trapezoid functions $T[a,b,c,d]$. 
This way, the degree to which a value fulfills a property can be computed with its membership function as follows:
\begin{equation}\label{eq:trap}
\mu_{T[a,b,c,d]}(x) = 
\begin{cases}
0,  & (x \leq a) \ or \ (x > d) \\
\frac{x-a}{b-a} &  a < x \leq b \\
1, &  b < x \leq c \\
\frac{d-x}{d-c} & c < x \leq d \\
\end{cases}
\end{equation}

\begin{table*}[t]
 \caption{Several types of fuzzy protoforms described in the LDD literature}
 \begin{tabular}{p{3.5cm}p{1.5cm}p{4cm}p{7cm}}
  \hline
 Authors                                          & Year & Protoform                                              & Verbalized example                                                              \\
  \hline
  
 Cariñena \cite{Carinena}                       &  1999& X was A in T                                           & Temperature was high in the last minutes                                        \\
                                                   && In T, X was A                                          & Shortly after the increase in pressure, temperature was high                    \\
 Castillo-Ortega et al. \cite{Castillo-OrtegaMS11} & 2001 & Q of D are A                                           & Most days of the cold season patient inflow was high                            \\
  
  Almeida et al.\cite{Almeida13}                  &  2013 & Q Y's are P $Q_t$ times                                & Most patients have high blood pressure most of the time.                        \\
                                                   && Q Y's with C are P                                     & Most patients with disease X have low blood pressure.                           \\
   Wilbik and Dijkman \cite{Dijkman17, Wilbik15}  & 2015, 2017 & In Q cases there was P                                 & In most cases there was a short throughput time                                  \\
                                                  && In Q cases, when condition R was fulfilled there was P & In most cases when "Registration" was short, there was a short throughput time. \\

  \hline
 \end{tabular}
 \label{tab:protoforms}
\end{table*}

Some limitations exist when using linguistic summaries of data.
In the literature mostly type-I and type-II protoforms are used without diving in a deeper natural language realization, however, presenting the user with a linguistic summary composed of multiple isolated type-I and type-II descriptions is not the most appropriate solution due to their lack of expressiveness and limited semantics\cite{Marin16, Ramos-Soto21}.
One direction which has been followed recently 
in order to improve this limited semantics is their extension with additional elements; as the temporal dimension (due to great availability of time series data) or other domain specific information.
Table \ref{tab:protoforms} includes some of the type-I and type-II protoforms that have been proposed in the literature on recent times that may be of inspiration in our case. On \cite{Marin16, Ramos-Soto21} a more extensive review of protoforms and applications can be found.

\subsection{Process Mining}
Process execution is recorded in event logs. 
Process mining goal is to exploit that recorded event data, by automatically discovering the underlying process model, to extract with it valuable, process related information in a meaningful way.
This information can be used to understand what is really happening in a process by providing insights
%, determining performance and detecting and identifying bottlenecks, 
which help to anticipate problems and streamline and improve processes \cite{Aalst16}.
Process mining serves as a bridge between classical business process model analysis and data mining or data science techniques.
On the one hand, classical process model analysis is a model-centric discipline; it puts all its emphasis on theoretic process models without giving much attention to the real execution data.
However, the value of a process model is limited if too little attention is paid to the alignment between the model and reality (recorded event data).
On the other hand, data mining techniques focus completely on the data without paying any attention to the model (or end-to-end processes).
These techniques are able to recall frequencies of events, number of events per case, and basic case statistics, but can not be used to analyze bottlenecks, expected behaviors, deviation, etc. so are not able to answer the most frequent questions when dealing with processes.
Current linguistic summarization techniques for process data \cite{Dijkman17, Wilbik15} focus solely on data mining techniques (they only use event log data) without paying attention to the underlying process model.
This makes evident the need to propose a new series of protoforms which do take in count both aspects of a process and are based on process-mining techniques.

%!TEX root = main.tex

In order to better describe the protoforms presented in this paper it is necessary to introduce some of the basic elements in a process and in an event log.

An activity $\alpha \in \cal{A}$, being $\cal{A}$ the set of all activities, is each well-defined step in a process.
Events $e$ represent the execution of an activity $\alpha$ in a particular instant.
They are characterized by two mandatory attributes: the executed activity $\alpha$ and the timestamp of the event; but they can have additional attributes such as its associated resources, their time duration, etc.

A trace is an ordered list of events where each event occurs at a certain instant relative to the other events in it i.e. it represents the sequence of events a case follows.
A case $c \in \cal{C}$, being $\cal{C}$ the set of all cases in the process, then represents a particular execution of the process and, as events, cases have attributes.
The most mandatory attributes of a case are its corresponding trace and its identifier.
Other attributes may be its throughput time, the customer involved in the case, the country of an order, etc. 
Table \ref{tab:log} shows an example of an event log, a multiset of cases $L=[\hat{c}_{1}, ..., \hat{c}_{n}]$.

\begin{table}[t]
 \caption{Event log example.}
 \centering
 \begin{tabular}{c c c c}
  \hline
  case\_id & event\_activity      & case\_sex & event\_time      \\
  \hline
  20629    & consultation         & Male      & 2013-06-04 09:00 \\
  20629    & special-consultation & Male      & 2012-06-14 09:00 \\
  20634    & echocardiogram       & Female    & 2012-06-21 09:00 \\
  20634    & consultation         & Female    & 2012-06-21 10:00 \\
  21657    & echocardiogram       & Male      & 2012-10-25 09:00 \\
  21657    & consultation         & Male      & 2012-10-25 10:00 \\
  \hline
 \end{tabular}
 \label{tab:log}
\end{table}

\begin{figure*}[!t]
 \centering
 \includegraphics[width=0.70\linewidth]{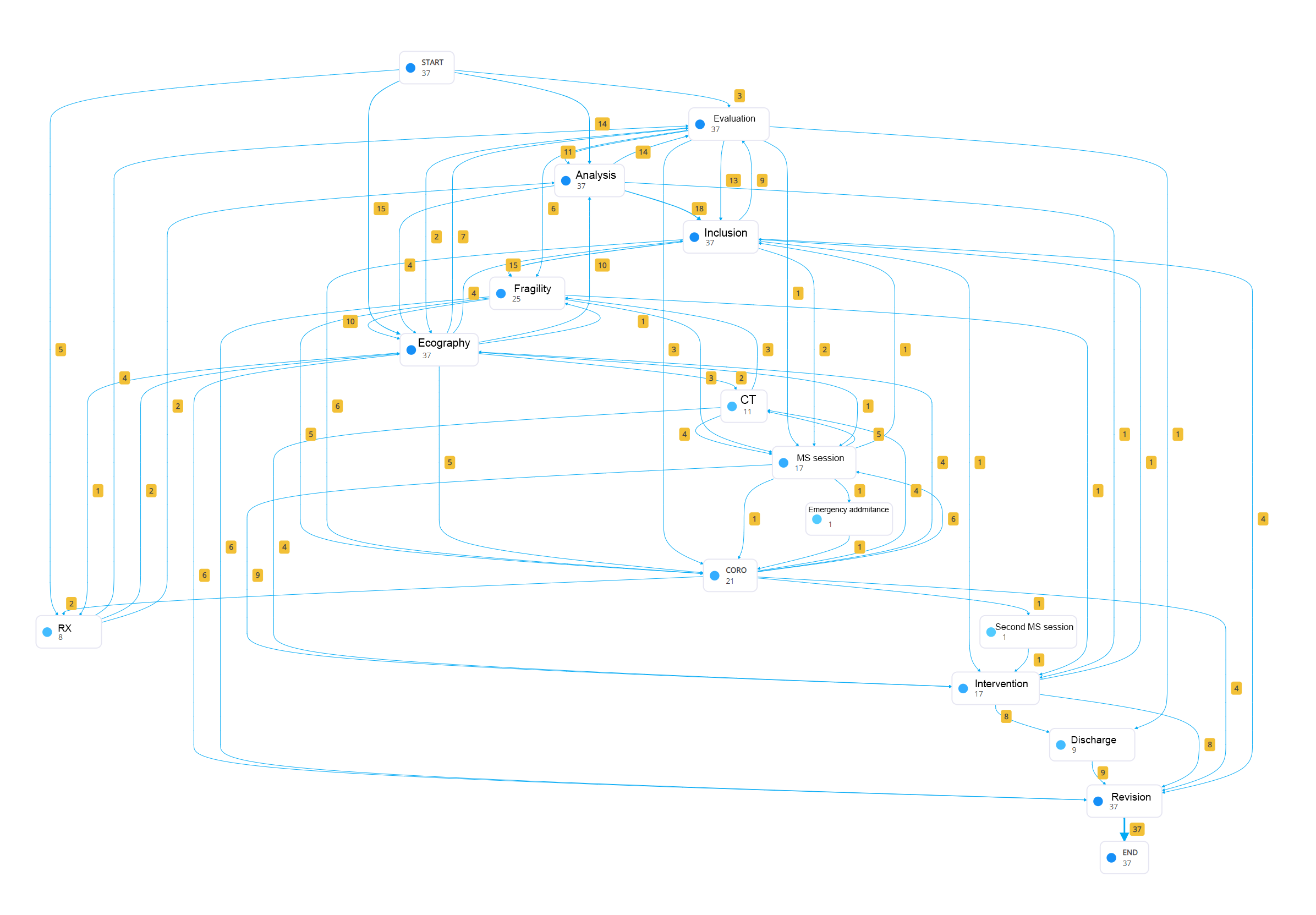}
 \caption{Simplified model of the valvulopathy process represented with the InVerbis Analytics visualization tool \cite{InVerbisAnalytics}.}
 \label{fig:process-model}
\end{figure*}

By applying discovery algorithms \cite{Aalst16} the model of a process can be extracted from an event log without any additional a-priori information.
The discovered model shows which activities take place (as nodes), its ordering and relations by describing causal dependencies between them (as arcs connecting the nodes). Figure \ref{fig:process-model} shows a simplified process model (only the top 0.03\% most common behavior is represented) of the use case here presented.

Establishing a relation between a process model and the event log the process model is extracted from is a key element in process mining.
This can be done by replaying \cite{Aalst16} an event log over its corresponding process model, and allows to exploit the four perspectives of process mining: organizational (information about resources), control-flow (ordering of activities), case (attributes of the cases) and time (timing and frequency of events) perspectives.
In this proposal we will focus mainly on the control-flow, case and time perspectives, since these are the perspectives where most questions are posed by experts in domains such as healthcare\cite{mans}. Putting special emphasis on the time perspective and providing ways to relate the three perspectives through new proposed protoforms.

%!TEX root = main.tex

\section{Proposal}\label{section:proposal}
In this section we propose a new series of protoforms for processes, using as a guide a case study in the health-care medical domain: the process related to the patients' management in the Valvulopathy Unit of the Cardiology Department of the University Hospital of Santiago de Compostela.
In this Unit, consultations and medical examinations, such as echocardiograms or Computed Tomography scans are performed to patients with aortic stenosis \cite{AOS} in order to decide their treatment (including surgery). 
Other information like unexpected events (e.g. non-programmed admissions) and patient management activities (e.g. inclusion in the process) are also recorded in the event log.

Medical experts show real interest in applying process mining techniques to this process, since it allows to extract valuable knowledge like, relationships between patients attributes (case attributes), relationships and delays between crucial activities (timing and frequency of events)
or different paths of the process patients with different attributes follow (control-flow perspective).
However, understanding this information is highly difficult for non process-expert users, thus why medical experts show interest in natural language descriptions of health-care medical process. 
The protoforms presented in this paper are derived from experts needs in order to fulfill their information requirements.

\subsection{Temporal contextualization of attributes}
This first protoform aims at describing how attributes of patients behave during different stages of the process e.g. "In year 2020, most patients had emergency admittance".
The extended protoform, also allows to see if any type of relation between attributes holds e.g. "In year 2020, most patients who underwent a surgery where older than 80".
We extend protoforms (\ref{eq:1}) and (\ref{eq:2}) to the following ones, respectively:
\begin{equation}\label{eq:3}
    In \ Ti, \ Q \ patients \ had \ attribute \ P 
\end{equation}
\begin{equation}\label{eq:4}
    In \ Ti, \ Q \ patients \ with \ attribute \ C \ had \ attribute \ P 
\end{equation}

Where $Q$ is a quantifier, $Ti$ is a time interval and $C$ and $P$ make reference to any of the attributes of a case that are present on the event log, or any additional attributes computed with expert knowledge after the process model is discovered. 
This process of computing new attributes can be seen as the Feature Engineering task of Machine Learning. 
Process mining techniques allow us to compute new attributes that domain experts show high interest on such as the waiting time between two particular activities, the number of times one activity triggers the execution of one another, etc. 

Attributes will be represented as linguistic variables to which the elements of the referential (patients or cases) will be evaluated against as in (\ref{eq:trap}).
In cases where the summarizer or qualifier are categorical, as well as for property $Ti$ (defined by its start and end dates) we propose the use of crisp linguistic values.
The membership of a value to these properties can be computed straightforward by taking $a=b$ and $c=d$ in expression (\ref{eq:trap}). 

Truth value of (\ref{eq:3}) is computed directly following \cite{Carinena}, furthermore, we propose an extension of it to calculate the truth value of (\ref{eq:4}) as:
\begin{equation} \label{eq:6}
 {T}  = \mu_Q \left( \frac{\sum_{i = 1}^n\mu_{Ti}(p_i) \wedge \mu_P(p_i) \wedge \mu_C (p_i) }{\sum_{i = 1}^n\mu_T(p_i) \wedge \mu_C(p_i)} \right)
\end{equation}

where $n$ is the number of patients (cases), $p$ represents a patient (case), $\wedge$ represents the t-norm minimum which is used as conjunction. In  (\ref{eq:4}) Zadeh's quantification model is used, although any valid quantification model \cite{DanielSanchez2013, Andrea20} could be considered.

\subsection{Causality and temporal relationship between events}
The time perspective is highly relevant in health-care medical processes, as wait times between activities can have a heavy impact on whether the treatment of a patient is successful or not.
Descriptions as "In year 2020, in most cases, patient evaluation takes place shortly after its inclusion" or in an extended form "In year 2020, in most cases where patients had emergency admittance, patient intervention takes place shortly after its MS session" can be generated thanks to the causal relationships between events that process mining is able to extract. 
Each event is characterized by its relationships with "input" and "output" events (i.e. events that happen consecutively before and consecutively after it) and the waiting time between their execution. 
These relationships are the ones described by the arcs in a process model \cite{Aalst16}.
Thus, a set of relationships and wait times between all activities in a process can be computed, allowing for the generation of the following proposed protoforms:
\begin{equation}\label{eq:9}
    In \ Ti, \ in \ Q \ cases \  R 
\end{equation}
\begin{equation}\label{eq:10}
    In \ Ti, \ in \ Q \ cases \ where \ patient \ had \ attribute \ C \  R 
\end{equation}

Where $Ti$, $Q$ and $C$ are as before and $R$ is a temporal relation between two activities $A$ and $B$ following the algebra proposed by Allen \cite{Allen83}. 
In this case, as the proposed process only has registered end timestamps, only precedence (after, before) relationships can be expressed. 
The relationship between two events $a$ and $b$ can be computed as:
\begin{equation}\label{eq:R}
r_{a,b}(x) = 
\begin{cases}
0,  & \text{activities not causally related} \\
T_a - T_b , &  \text{activities are causally related} \\
\end{cases}
\end{equation}

So $r_{a,b}$ is zero if events are not causally related, positive if origin event precedes destination event and negative if destination event precedes origin event.
$R$ can be computed for each pair of activities (for all the executions of each activity) in the process (in both directions $A$ before $B$, $B$ before $A$).
This way, linguistic variable "after" could be described as a series of positive, non monotonous fuzzy sets after = [immediately after, shortly after, after, long after, at some point after]. And "before" in a similar way but with negative, non monotonous sets.
This makes for a similar truth evaluation process as before, where truth degree for (\ref{eq:9}) is computed following \cite{Carinena} and for (\ref{eq:10}) with (\ref{eq:6}) substituting $P$ for $R$ in both cases.

These descriptions give an easy understanding of the behaviour and different paths (control-flow perspective) patients follow in the process, plus, the addition of the wait time between events (time perspective) allows medical experts to determine whether the behavior of the process is being as expected, where excessive wait times are happening and how activities relate among them.

\subsection{Deviance protoforms}
These protoforms aim at putting in relevance attributes which may be causing deviance over other attributes. 
For example "In year 2019, most patients had a normal waiting time between the MS session of the patient and its intervention. However, several patients with emergency admittance had a short waiting time". 
This protoform is indeed a composite protoform, obtained by composing two protoforms: a type-I general summary protoform with a type-II contrasting protoform, through a semantic relation 
%which has an associated relation strength 
\cite{Ramos-Soto21}.
And it is defined as:
\begin{equation} \label{eq:7}
\begin{aligned}
In \ T, \ Q_1 \ patients \ had \ attribute \ P_1. \ However, \\ Q_2 \ patients \ with \ attribute \ C \ had \ attribute \ P_2
\end{aligned}
\end{equation}
A particular case of this protoform is one in which the first statement is made over the expected value of some attribute, defined by experts i.e. "In year 2020, the waiting time between the MS session of a patient and its intervention is expected to be around 25 days. However, most patients from ambulatory admittance had a longer waiting time":
\begin{equation} \label{eq:8}
\begin{aligned}
In \ T, \ P_1 \ is \ expected. \
However, \\ Q_2 \ patients\  with \ attribute \ C \ had \ attribute \ P_2
\end{aligned}
\end{equation}

For these protoforms, the truth value may be derived from the aggregation of the truth values of its constituents through any t-norm. For simplicity and consistency we propose the use of the t-norm minimum.
If we refer to the general protoform as $S1$ and to the contrasting protoform as $S2$ the truth value of the deviance protoform (\ref{eq:7}) can be computed as:
\begin{equation}
    T = T(S1) \wedge T(S2)
\end{equation}
where $T(S1)$ and $T(S2)$ can be computed with (\ref{eq:3}) and (\ref{eq:4}).
For protoform (\ref{eq:8}) where $P1$ is the expected value defined by the experts, there is no need to assess the truth degree of $S1$, as it will be always maximal. In this case, the truth degree of the composite protoform is $T(S2)$.

Other protoform combinations are possible.
Behavior deviance descriptions like "In year 2020, in most cases, the intervention of a patient takes place shortly after its MS session. However, most patients with a low number of medical tests a second MS session takes place after the first MS session is performed" are of high interest for medical experts, as they allow them to detect bottlenecks and unexpected behaviours that would otherwise remain unknown.
Also, type-II protoforms could be used for both protoforms, allowing to compare different categories of patients for some attribute, e.g. "In year 2020, most male patients had a short waiting time between the MS session of the patient and its intervention. However, most female patients had a normal waiting time between the MS session of the patient and its intervention".
These descriptions allow medical experts to easily grasp if differences between groups of patients exist for certain attribute; as it may be their sex, type of admittance, treatment patient is being given, etc.

\subsection{Generation Pipeline}
On the one hand, in LDD approaches,
linguistic summaries are generated by a search (exhaustive or non-exhaustive) through the semantic space; the power set of all protoform instances that can be built using the defined quantifiers, qualifiers and summarizers guided by quality measures (truth value, strength of relation, etc.).
On the other hand, D2T and NLG systems, as our proposal, follow a pipeline where the main stages of the generation process related to handling of data (data interpretation and document planning) use expert knowledge to determine which messages must be included and realized into the final text. This expert knowledge usually takes the form of sets of rules, but other approaches as machine learning, or statistical tests which in our case are used to determine whether a deviance protoform may present relevant information to the user, can be used.

%!TEX root = main.tex

\section{Validation and Conclusions }\label{sec:conclusions}
In this section we present the assessment of the proposed model in a real domain of application (activities and attributes of the patients of a cardiac valvulopathy unit). 
The assessment was conducted by medical experts of the Cardiology Department, University Hospital of Santiago de Compostela through a questionnaire, created by taking general ideas of the Technology Acceptance Model (TAM)\cite{TAM} adapted to linguistic summarization that was used in \cite{Dijkman17}.  

Instances of each type of protoform are presented to the experts to assess the degree to which protoforms provide useful information in a comprehensible way (examples can be found in previous sections).
Furthermore, the process model is also included to assess whether natural language descriptions are preferable or not to graphical representations.
Finally, general questions are asked in order to determine
f the use of natural language descriptions of processes in the health-care domain is found useful.
All questions are asked on a five-level Likert scale ranging from 'strongly disagree' to 'strongly agree'. 

Results show most protoforms are found to provide interesting information, except those cases where routine information was given.
However, by proposing the same protoforms with different data, we were able to recall that the perceived usefulness is only found lower when protoforms convey information medical experts already know and not because the proposed protoforms were not correct. 
When the information conveyed in the description is data unknown to medical experts, descriptions were labelled as really interesting.
All protoforms, except one case where the realization was not suitable, were found comprehensible and easy to understand.
In all cases where natural language descriptions were confronted with graphical representations, a clear preference was shown for natural language descriptions.
From the general questions, medical experts clearly stated that natural language descriptions are useful, give them a better understanding of what is happening in the process, allow them to complete tasks quicker, increase the quality of their work and increase their effectiveness.

In this paper we present an approach to obtain natural language descriptions of health-care processes. 
We propose a series of protoforms which include temporal and causal information from processes as well as patient attributes, that are able to quantify attributes in time during a process life-span, recall causal relations and temporal distances between events, and describe whether differences exist in attributes between different groups of patients. 
By introducing the temporal dimension through imprecise descriptions of frequency and temporal characteristics of attributes and activities and through the composition of protoforms, the semantics and expressiveness of our proposal is greatly enhanced.
We propose to generate the descriptions using a novel approach based the D2T pipeline using process-mining techniques and expert knowledge.
A real health-care use case is presented, showing the potential of the proposed protoforms for providing natural language descriptions addressed to cardiology specialists about activities and attributes of the patients of a cardiac valvulopathy unit.
%!TEX root = main.tex

%%%%%%%%%%%%%%%%%%%%%
\section*{Acknowledgments}
\label{sec:acks}
%%%%%%%%%%%%%%%%%%%%%

Thanks to Dr. Carlos Peña and Dr. Violeta González from the Department of Cardiology, University Clinical Hospital of Santiago de Compostela, SERGAS, Biomedical Research Center in the Cardiovascular Diseases Network (CIBER-CV), for providing the anonymized data and validating the proposal.

\bibliographystyle{IEEEtran}
\bibliography{ref}

\end{document}